\documentclass[letterpaper, 10 pt, conference, twoside]{ieeeconf}  

\IEEEoverridecommandlockouts                              

\usepackage{graphics} 
\usepackage{mathptmx} 
\usepackage{times} 

\usepackage{bm}
\usepackage{dsfont}
\usepackage{subcaption}
\usepackage[hyphens]{url}
\usepackage[mathscr]{euscript}
\usepackage{booktabs}
\usepackage[normalem]{ulem}
\usepackage{xcolor}
\usepackage[T1]{fontenc}
\usepackage[utf8]{inputenc}
\usepackage{mathtools}
\usepackage[bookmarks=true]{hyperref}
\usepackage{ stmaryrd }
\graphicspath{{./Graphics/}}

\usepackage{amsmath, amssymb}
\usepackage{enumerate}
\usepackage{tikz}
\usepackage{multirow}

\usetikzlibrary{shapes.geometric}
\newcommand{\myignore}[1]{}

\def \F {\mathbf{F}}
\def \G {\mathbf{G}}

\def \U {\mathbf{U}}

\def \X {\mathbf{X}}

\def \SpecMDP {\mathscr{M}_{Spec}}
\def \SpecFSM {\mathscr{M}_{\{\varphi\}}}

\DeclareMathOperator*{\argmin}{arg\,min}
\def \trace {[\bm{\alpha}]}
\def \lab {\mathscr{L}(\trace)}
\def \envMDP {\mathscr{M}_{\mathscr{X}}}
\def \FSMStates {\{ \langle \varphi' \rangle \}}

\title{\LARGE \bf
Interactive Robot Training for Non-Markov Tasks
}

\author{Ankit Shah$^{1}$, Samir Wadhwania$^{2}$, Julie Shah$^{3}$
\thanks{$^{1}$Ankit Shah is a PhD candidate at the Computer Science and Artificial Intelligence Laboratory at the Massachusetts Institute of Technology.
        {\tt\small ajshah@mit.edu}}%
\thanks{$^{2}$Samir is a PhD student Computer Science and Artificial Intelligence Laboratory at the Massachusetts Institute of Technology.
        {\tt\small samirw@mit.edu}}%
\thanks{$^{3}$Julie Shah is an associate professor at the Massachusetts Institute of Technology}
}

\begin{document}

\maketitle
\thispagestyle{empty}
\pagestyle{empty}

\begin{abstract}
Defining sound and complete specifications for robots using formal languages is challenging, while learning formal specifications directly from demonstrations can lead to over-constrained task policies. In this paper, we propose a Bayesian interactive robot training framework that allows the robot to learn from both demonstrations provided by a teacher, and that teacher's assessments of the robot's task executions. We also present an active learning approach -- inspired by uncertainty sampling -- to identify the task execution with the most uncertain degree of acceptability. Through a simulated experiment, we demonstrate that our active learning approach identifies a teacher's intended task specification with an equivalent or greater similarity when compared to an approach that learns purely from demonstrations. Finally, we demonstrate the efficacy of our approach in a real-world setting through a user-study based on teaching a robot to set a dinner table.

\end{abstract}

\section{Introduction}

Humans are adept at quickly learning to perform multi-step tasks like setting a dinner table, clearing a desk, or assembling furniture. Tasks such as these typically involve temporal elements like adherence to constraints or decomposition into and prioritization of sub-tasks. Linear temporal logic (LTL) \cite{pnueli1977temporal} provides an expressive grammar for modeling a range of such non-Markov temporal properties; however, formal languages like LTL are often unwieldy for the average user. In order to facilitate rapid deployment of capable robots to novel scenarios and tasks, it is desirable to allow users with task-specific expertise to directly program robots.

There has been a considerable amount of research related to inferring formal specifications through intuitive interfaces such as demonstrations \cite{shah2018bayesian, kim2019ijcai} and preferences expressed as natural language instructions \cite{oh2019planning, gopalan2018sequence}. To resolve the ambiguity associated with these teaching modalities, we proposed planning with uncertain specifications (PUnS) \cite{shah2019planning}, a framework for generating task plans wherein specifications are expressed as a belief ($P(\varphi)$) over LTL formulas. However, policies computed to optimize the PUnS criteria generate task executions that attempt to satisfy a large number of candidate formulas, potentially over-constraining task execution. In this paper, we demonstrate that belief over LTL formulas can also serve to identify task executions with an uncertain degree of acceptability . These executions can then be demonstrated back to the user to elicit an assessment of their acceptability, which in turn can reduce the uncertainty of the distribution. 

We also propose an active querying strategy for identifying and performing such ambiguous task executions, and evaluate the performance of this active learning approach compared with learning purely from demonstrations (termed \textit{Batch}) and another interactive approach wherein task executions are generated by performance of random actions (termed \textit{Random}). Through results obtained from a simulation experiment, we demonstrate that our proposed method yields posterior belief distributions with higher similarity  to the ground truth specification as compared to \textit{Batch} and \textit{Random} approaches for a wide range of ground truth task specifications. Finally, we conducted a user study involving training a robot to set a dinner table using our active learning approach, with demonstrations provided either in-person or by remotely operating  the robot. Our findings indicate the efficacy of our active learning approach for learning task specifications that are well aligned with the ground truth specifications (average similarity: $0.86~ 95\%~ \text{CI } [0.82, 0.92]$).

\section{Related Work}
The objective of allowing domain experts to directly program robots has driven research into methods for programming through intuitive modalities. Prior research has yielded models for learning a teacher's intended task by processing input provided by the teacher  through demonstrations \cite{argall2009survey,chernova2014robot}, natural language instructions \cite{luketina2019,oh2019planning,gopalan2018sequence}, corrections \cite{bajcsy2017learning,bajcsy2018}, or preferences \cite{dorsa2017active, biyik2018batch, biyik2019asking}. One key feature of our approach is the ability to model temporal tasks by using LTL as the specification language. Chanlette-Vazquez et al . \cite{vazquez2018} proposed observing the demonstrated task execution given the true specification as a maximum entropy estimator . Kasenburg and Scheutz \cite{kasenberg2017interpretable} proposed an optimization-based framework for modeling a decision maker's behavior as an LTL formula. Camacho et al. \cite{camacho2019learning} developed an exact method for mining LTL formulas based on sets of satisfying and non-satisfying traces for the shortest LTL-F(inite) formula. Shah et al. \cite{shah2018bayesian} proposed a Bayesian approach to specification inference to model the uncertainty associated with inferring task specifications from a small number of  demonstrations. While most of the previous work on learning non-Markov task specifications has focused on learning  solely from teacher's demonstrations, in this paper, we adopt an iterative Bayesian approach that unifies  the teacher's input provided via demonstrations or as assessments of the robot's task executions.

There has also been considerable interest in developing algorithms that allow the learner to elicit the teacher's feedback (an active learning paradigm). One expected benefit of an active approach is that the learner can guide the teacher's feedback such that it is optimally impactful to the learner’s own behavior. Cakmak et al. \cite{cakmak2012designing, cakmak2010designing} developed a taxonomy of queries that allow a learner to refine its understanding of the task specifications. Sadigh et al. \cite{dorsa2017active} proposed an active learning framework for sequential decision-making problems that relies upon pairwise preference between candidate trajectories selected according to a maximum volume removal heurisitc; Biyik et al. \cite{biyik2019asking} then extended this to generate queries using maximum information gain criterion. Biyik and Sadigh \cite{biyik2018batch} proposed a batch active framework for preference-based learning wherein multiple queries are generated simultaneously instead of one at a time. Cui and Niekum \cite{cui2018active} proposed an active learning model based on information gain that operates on individual state-action pairs, allowing segments of the trajectory to be labeled ``desirable” and ``undesirable.” However, present research into active learning for robotics has largely focused on formulations that represent the underlying task as a Markov decision process (MDP), with the state space known a priori. 

Admitting non-Markov task specifications would increase the robot’s ability to handle complex tasks . Therefore, prior research has led to development of planning algorithms for hybrid controller synthesis \cite{kress2009temporal}, symbolic planning \cite{camacho2018finite, camacho2019strong}, and reinforcement learning \cite{littman2017environment,toro2018teaching, icarte2018using, camacho2019ltl}. In this paper, we build upon  planning with uncertain specification (PUnS) \cite{shah2019planning}, a problem formulation that allows task specifications to be expressed as a belief over multiple LTL formulas. Policies computed to optimize the PUnS evaluation criteria satisfy the entire belief distribution rather than a single LTL formula, allowing the learner to reconcile the ambiguity inherent in the teacher's demonstrations. Our proposed extension leverages the reward machine \cite{camacho2019ltl} representing the learner's belief over LTL formula in order to identify a task execution suitable for active learning.

\begin{figure*}
    \centering
    \includegraphics[width = 0.9\textwidth]{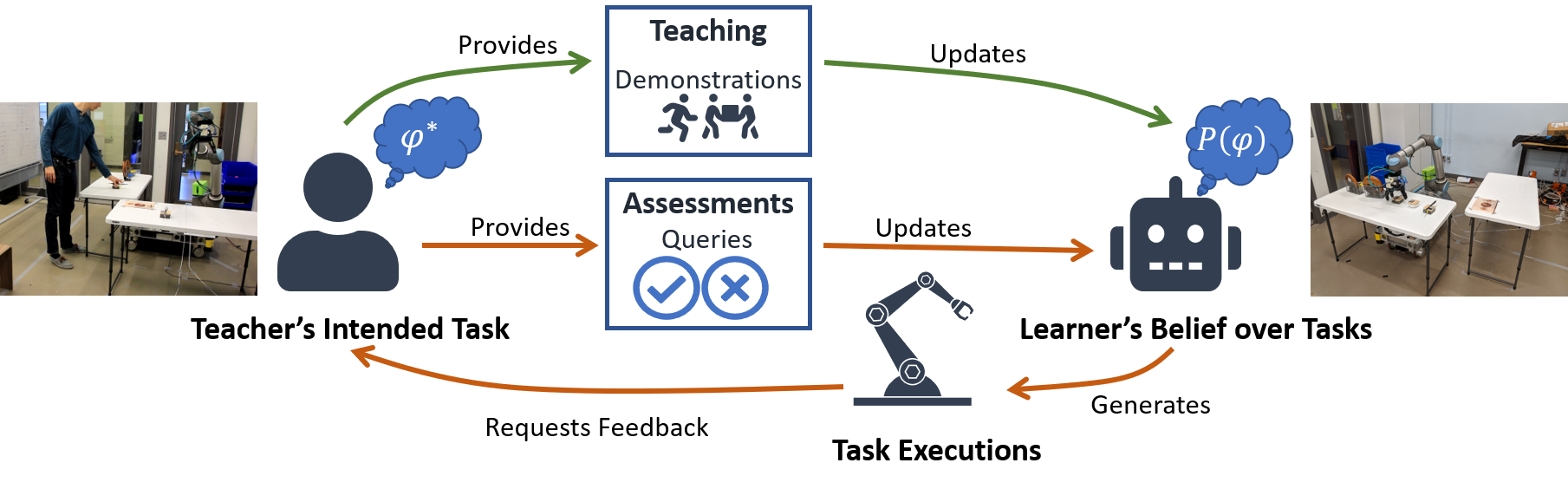}
    \caption{Our proposed Bayesian interactive learning framework that unifies learning from demonstrations provided by the teacher and using informative queries generated by the learner to refine the learner's belief. The green path depicts the teacher initiating training using task demonstrations. The orange path indicates the learner initiating training by demonstrating a task execution as a query requesting an assessment from the teacher.}
    \label{fig:framework}
    \vspace{-15 pt}
\end{figure*}

Our contribution in this paper is twofold. First, we propose a novel interactive learning framework (Figure \ref{fig:framework}) for non-Markov tasks that unifies teacher inputs through demonstrations and assessments of the learner's task execution . Second, we develop an active learning approach that leverages the reward machine representation of an instance of a PUnS problem to identify task executions with the most uncertain degree of acceptability.

\section{Preliminaries}
\label{sec:prelim}

\subsection{Linear Temporal Logic}
Linear temporal logic (LTL), first proposed by Pnueli \cite{pnueli1977temporal}, provides a flexible grammar for defining temporal properties over Boolean propositions. A valid LTL formula is constructed using atomic propositions (discrete time sequences of Boolean values) and logical and temporal operators. The truth value of an LTL formula is evaluated for traces $[ \bm{\alpha} ]$ for a set of atomic propositions, $\bm{\alpha}$. The notation $[ \bm{\alpha}], t \models \varphi$ indicates that formula $\varphi$ holds at time $t$. Trace $[ \bm{\alpha} ]$ satisfies $\varphi$ (denoted as $[ \bm{\alpha} ] \models \varphi$ iff $ [ \bm{\alpha} ] ,0 \models \varphi$. The minimal syntax of LTL is as follows:

\begin{equation}
  \varphi::= p \mid \neg\varphi_1 \mid \varphi_1\vee\varphi_2 \mid \X \varphi_1 \mid \varphi_1\U\varphi_2
  \label{Eq:LTLsyntax}
\end{equation}

Here, $p$ is an atomic proposition, and $\varphi_1$ and $\varphi_2$ represent valid LTL formulas. The operator $\X$ is read as ``next'' and $\X \varphi_1$ evaluates as true at $t$ if $\varphi_1$ holds at $t+1$. The operator $\U$ is read as ``until'' and $\varphi_1 \U \varphi_2$ evaluates as true at $t$ if $\varphi_2$ holds at some time $t_2 > t_1$ and $\varphi_1$ holds for all $t$, where $t_1 \leq t \leq t_2$. In addition to the minimal syntax, we also incorporate the conjunction operator $\wedge$, along with two other temporal operators: $\F$ (eventually) and $\G$ (globally). $\F \varphi_1$ holds at $t_1$ if $\varphi_1$ holds for some time $t \geq t_1$; similarly, $\G \varphi_1$ holds at $t_1$ if $\varphi_1$ holds for all $t \geq t_1$.

Finally, a progression $\text{Prog}(\varphi, \alpha_t)$ over an LTL formula with respect to the truth assignment, $\alpha_t$, is defined such that $ \forall \trace: ~ [\alpha_t, [\bm{\alpha}]],t \models \varphi ~ \text{iff} ~ \trace,t+1\models \text{Prog}(\varphi,\alpha_t)$. A progression of an LTL formula with respect to a particular truth assignment must hold at the next time step in order for the original formula to hold at the current time step. We use the syntactic progression rules defined by Bacchus and Kabanza \cite{bacchus2000using} to compute formula progressions.


\subsection{Markov Decision Process}
A Markov decision process (MDP) is a planning problem defined as a tuple $\mathscr{M} = \langle \mathscr{S}, \mathscr{A}, T, R \rangle$, where $\mathscr{S}$ represents the set of all possible states; $\mathscr{A}$ is the set of actions available to the learner; $T \coloneqq P(s' \mid s, a)$ is a probability distribution over the next state $s' \in \mathscr{S}$ given current state $s\in S$, and the action $a \in A$ executed at the current time step; and $R : \mathscr{S} \rightarrow \mathbb{R}$ is the reward function that returns a scalar value given the current state. 



\section{Interactive Training for Non-Markov Tasks}

\subsection{Problem Formulation}

In this setting, the teacher intends to teach  a task represented by an LTL formula, $\varphi^*$, unknown to the learner. The learner maintains a belief over candidate LTL formulas $P(\varphi)$; this distribution is defined as a mass function, $P:\bm{\varphi} \rightarrow [0,1]$. The support of $P(\varphi)$ is restricted to a discrete set of LTL formulas, $\{\varphi\}$, where each formula represents a property belonging to the ``Obligations" class as defined by Manna and Pnueli \cite{manna1987hierarchy}. The learner's degree of success is determined by comparing the similarity between the belief distribution and the ground truth LTL formula (Equation \ref{eq:sim}).

The learner represents the task environment as a state, $x \in \mathscr{X}$, and also has access to a set of actions, $\mathscr{A}$. The state of the system, $x$, maps to a set of finite known Boolean propositions, $\bm{\alpha}\in \{ 0,1\}^{n_{prop}}$, through a labeling function, $f: \mathscr{X} \rightarrow \{ 0,1 \}^{n_{prop}}$. We assume that a trace of propositions, depicted by $\trace$, is sufficient to determine the truth value of any formula within the support $\{ \varphi \}$ of the learner's belief; thus, any task execution, whether generated by the robot or demonstrated by the teacher, is represented as a trace, $\trace$. We also define a Boolean label, $\mathscr{L}(\trace) \in \{0,1\}$, that indicates whether the given trace is acceptable. For the purposes of this paper, we assume that all task executions demonstrated by the teacher are labeled as acceptable, and that the teacher's assessment of the executions demonstrated by the learner is perfect.

\subsection{Overview of the Interactive Learning Framework}

Figure \ref{fig:framework} depicts our proposed interactive framework for training a learner using a combination of demonstrations provided by a teacher and that teacher's assessments of task executions generated as a query by the leaner . The learner must carry out two processes: \textit{learning}, wherein the robot updates its belief conditioned upon labeled task executions; and \textit{planning}, where it must use that belief to generate task executions. We adopt an iterative version of our prior work on Bayesian specification inference \cite{shah2018bayesian}, and extend it to allow both positive and negative examples (as elaborated upon in Section \ref{ss:bsi}). Formally, if the learner's initial belief over formulas is $P^{i}(\varphi)$ and the learner receives a dataset of task executions and their labels, $\mathscr{D} = \{ \langle \trace, \mathscr{L}(\trace) \rangle \}$, then the learner computes an estimate of the posterior distribution, $P(\varphi \mid \mathscr{D})$. The learner updates its belief to be the computed posterior as follows:

\begin{equation}
    P^{i+1}(\varphi) \gets P(\varphi \mid \mathscr{D})
    \label{eq:update}
\end{equation}

The learner has the ability to compute two types of policies depending upon the availability of a teacher to assess its task executions. If an assessment is unavailable, the learner computes a policy to satisfy its current belief, $P^i(\varphi)$. (This is an instance of planning with uncertain specifications (PUnS) \cite{shah2019planning}, as briefly described in \ref{ss:puns}.) The original non-Markov planning problem is compiled into an equivalent MDP representation, with a reward function representing the \textit{minimum regret} criterion \cite{shah2019planning}.

If a teacher's assessment  is available, the learner computes a policy to generate a task execution with the most uncertain degree of acceptability as per the learners current belief, $P^i(\varphi)$. The teacher's assessment of this task execution would be most beneficial for reducing the learner's uncertainty of the true specification; we describe our approach to generating such an informative query in Section \ref{ss:query}.

\subsection{Bayesian Specification Inference}
\label{ss:bsi}
Bayesian specification inference \cite{shah2018bayesian} is a probabilistic model for using demonstrations provided by a teacher to infer LTL formulas corresponding to the task specifications. According to this model, the hypothesis space of candidate LTL formulas comprises the set of formulas corresponding to the following template, which includes conjunctions of temporal behaviors identified by Dwyer et al. \cite{dwyer1999patterns}:

\begin{equation}
\varphi = \varphi_{global} \wedge \varphi_{eventual} \wedge \varphi_{order}
\label{eq:template}
\end{equation}

In our previous work \cite{shah2018bayesian}, we also proposed a domain-independent approximation of the likelihood function $P(\trace \mid \varphi)$ — depending upon the number of conjunctive clauses — that satisfied the size principle \cite{tenenbaum2000rules}. (A restrictive hypothesis has greater likelihood  than a less-restrictive hypothesis in the presence of data conforming to both.) Our approach is founded upon the classical interpretation of probability championed by Laplace \cite{laplace1951philosophical}, which involves computing the probabilities in terms of equally likely outcomes. If $N_{conj}$ conjunctive clauses exist within a formula, $\varphi$, there are $2^{N_{conj}}$ possible outcomes in terms of the truth values of the conjunctive clauses. In the absence of additional information, we  assign equal probabilities to each of the potential outcomes. Consider two candidate formulas, $\varphi_1$ and $\varphi_2$, with $N_{conj_1}$ and $N_{conj_{2}}$ conjunctive clauses and $\trace \models \varphi_1$. If this trace is considered acceptable (($\mathscr{L}(\trace) = 1$), the approximate likelihood odds ratio is computed as follows:

\begin{equation}
  \frac{P(\langle \trace, \lab = 1 \rangle \mid \varphi_1)}{P(\langle \trace, \lab = 1 \rangle \mid \varphi_2)} =
  \begin{cases}
  \frac{2^{N_{conj_1} } }{2^{N_{conj_2}}}
   &, [\bm{\alpha}] \models \varphi_2 \\
  \frac{2^{N_{conj_1}}}{\epsilon}
   &, [\bm{\alpha}] \nvDash \varphi_2
\end{cases}
  \label{er:positive}
\end{equation}

If trace $\trace$ is labeled as unacceptable ($\lab = 0$) and $\trace \nvDash \varphi_1$, the likelihood odds ratio is computed following the classical probability interpretation as before . With $2^{N_{conj}}$ conjunctive clauses, there are $2^{N_{conj}}-1$ possible evaluations of each of the individual clauses that would result in the given trace not satisfying the candidate formula; thus, the likelihood odds ratio is computed as follows:

\begin{equation}
    \frac{P(\langle \trace, \lab = 1 \rangle \mid \varphi_1)}{P(\langle \trace, \lab = 1 \rangle \mid \varphi_2)} =
  \begin{cases}
  \frac{2^{N_{conj_1}} (2^{N_{conj_2}} - 1)}{2^{N_{conj_2}} (2^{N_{conj_1}} - 1)}
   &, [\bm{\alpha}] \nvDash \varphi_2 \\
  \frac{2^{N_{conj_1}}}{\epsilon (2^{N_{conj_1}} - 1)}
   &, [\bm{\alpha}] \models \varphi_2
\end{cases}
\end{equation}

We assume that each data point in a given dataset $\mathscr{D} = \{ \langle \trace, \lab \rangle \}$ is independent of the others; thus, the likelihood of the entire dataset is the product of the individual likelihoods, as follows:
\begin{equation}
    P(\mathscr{D} \mid \varphi) = \prod_{\langle \trace_i, \lab_i \rangle \in \mathscr{D}} P(\langle \trace_i, \lab_i \rangle \mid \varphi)
\end{equation}

The probabilistic model is implemented in webppl \cite{dippl}, a universal probabilistic programming language. The posterior is approximated using webppl's Markov chain Monte Carlo algorithm with the Metropolis-Hastings acceptance criterion.

\subsection{Planning with Uncertain Specifications}
\label{ss:puns}
Planning with uncertain specifications (PUnS) \cite{shah2019planning} is a formulation for planning problems wherein task specifications are known as beliefs over LTL formulas, $P(\varphi)$. An instance of a PUnS problem is defined by the planning environment, which is encoded as an MDP sans a reward function, $\mathscr{M}_{\mathscr{X}} = \langle \mathscr{X}, \mathscr{A}, T_{\mathscr{X}} \rangle$; a task specification represented as a belief over LTL formulas, $P(\varphi)$, with support over a finite set of formulas, $\{ \varphi \}$; and one of the four evaluation criteria  proposed by us (Shah et al. \cite{shah2019planning}) for satisfying a belief over LTL formulas.

Consider a planning domain representing the task of setting a dinner table, with three objects accessible to the robot: a fork, a bowl, and a plate. The environment MDP, $\envMDP$, consists of a discrete state space, $\mathscr{X}$, that encodes whether each object is correctly placed; a discrete action space, $\mathscr{A}$, that encodes the selection of the object to be placed next; and the transition function, $T_{\mathscr{X}}$, which encodes how the action selection affects a change in the state. The acceptability of the task execution is evaluated using the vector of Boolean propositions, $\bm{\alpha} = [Fork,~ Bowl,~ Plate]$, where each proposition represents whether that object was correctly placed on the table. An example of a belief over the task specifications is represented by the distribution $P(\varphi)$, whose support, $\{ \varphi \} = \{\varphi_1 = \G ~\neg Fork \wedge \F~ Bowl \wedge ~\neg Bowl ~\U~Plate, ~ \varphi_2 = \G~ \neg Fork \wedge \F ~ Bowl\}$. The probabilities are as follows: $P(\varphi_1) = 0.3$, and $P(\varphi_2) = 0.7$. $\varphi_1$ encodes the specification  that the fork must never be placed, the bowl must be placed eventually, and that the bowl must not be placed until the plate has been placed; $\varphi_2$ encodes that the form must never be placed, and that the bowl must be placed eventually. Thus, any task execution that satisfies $\varphi_2$ also satisfies $\varphi_1$; however, the converse is not true. In order to perform the task to best align with this belief over specifications, one must place the plate, then the bowl, and must not place the fork. The PUnS formulation \cite{shah2019planning} formalizes this intuition.  

\begin{figure}
    \centering
    \includegraphics[width = 0.45\textwidth]{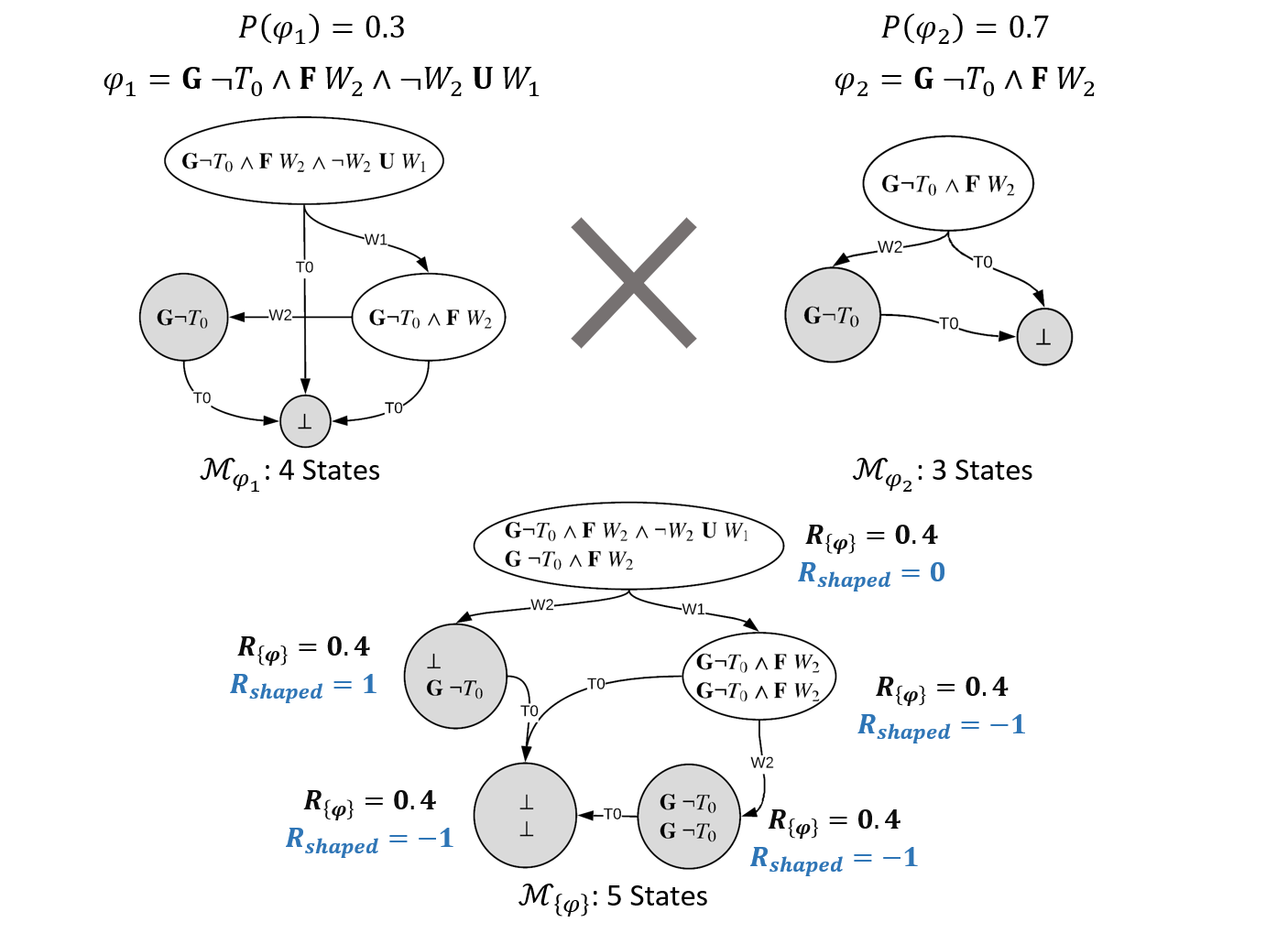}
    \caption{Example compilation process with $\{ \varphi \} = \{ \varphi_1, \varphi_2\}$ and the \textit{minimum regret} criterion. The deterministic MDPs $\mathscr{M}_{\varphi_1}$, and $\mathscr{M}_{\varphi_2}$ are composed through a cross product to yield the deterministic MDP $\SpecFSM$ corresponding to the set $\{ \varphi \}$. The reward based on the \textit{minimum regret} criterion ($R_{\{\varphi\}}$) is indicated in black, while the value of the shaped reward function ($R_{shaped}$) that enables the most uncertain task execution is indicated in blue.}
    \label{fig:example}
\end{figure}

In order to compute the policy to satisfy an instance of PUnS, we  first compile the non-Markov belief $P(\varphi)$ into an equivalent deterministic MDP $\SpecFSM$. The graphical representation of the compilation process for the dinner table example is depicted in Figure \ref{fig:example}. Formally, $\SpecFSM = \langle \FSMStates, \{0,1\}^{n_{prop}}, T_{\{ \varphi \}}, R_{\{\varphi\}}  \rangle$. $\FSMStates$ is the set of ordered tuples $\langle \varphi' \rangle$ that represent all progressions of the formulas contained in $\{ \varphi \}$, and the actions represent the truth values of the propositions, $\bm{\alpha}$. Let $\varphi'^{i}$ represent the $i^{th}$ formulas in the tuple $\langle \varphi' \rangle$; the transition function $T_{\{\varphi\}}$ is then defined as follows:

\begin{equation}
  T_{\{ \varphi \}}(\langle \varphi'_1 \rangle, \langle \varphi'_2 \rangle , \alpha) = \begin{cases}
                                      1, & \mathrm{if }~ \varphi'^i_2 = \mathrm{Prog}(\varphi'^i_1, \alpha) ~\forall~ i\\
                                      0, & \mathrm{otherwise}
                                    \end{cases}
\end{equation}

Let $\langle \bm{\varphi}' \rangle_{term}$ be the set of terminal states, where each of the component formula has either been satisfied ($\top$), dissatisfied ($\bot$), or has progressed to a safe-LTL formula. The reward function depends upon the choice of the PUnS evaluation criterion. The \textit{minimum regret} criterion is linearly dependent on the probability of the task execution being acceptable as determined by the belief $P(\varphi)$  For the \textit{minimum regret} criterion, the reward function is defined as follows:

\begin{equation}
    R_{\{\varphi\}}(\langle \varphi' \rangle) = \begin{cases}
    \sum_{i} P(\varphi^i)r(\varphi'^i) &,~ \text{if}~ \langle \varphi' \rangle \in \langle \bm{\varphi}' \rangle_{term}\\
    0 &,~ \text{otherwise}
    \end{cases}
\end{equation}

where $r(\varphi'^i)$ is defined as follows:
\begin{equation}
    r(\varphi'^i) = \begin{cases}
    1 &,~ \varphi'^i = \top \text{ or } \varphi'^i \in \text{ safe-LTL}\\
    -1 &,~ \varphi'^i = \bot\\
    \end{cases}
\end{equation}




This compiled deterministic MDP, $\SpecFSM$ is then composed with $\envMDP$ to obtain an MDP equivalent of the original PUnS problem, defined as follows:
\begin{equation}
    \mathscr{M}_{spec} = \langle \mathscr{X}\times \FSMStates, \mathscr{A}, T_{spec}, R_{\{\varphi\}} \rangle
\end{equation}

Here, 

\begin{equation}
  \begin{split}
  T_{Spec}(\langle \langle\varphi'_1 \rangle, x_1 \rangle, \langle \langle \varphi'_2\rangle,x_2 \rangle, a) &= \\
  T_{\{\varphi\}}(\langle \varphi'_1 \rangle, \langle \varphi'_2 \rangle, &f(x_2)) \times
  T_{\bm{\mathscr{X}}}(x_1, x_2, a)
\end{split}
\end{equation}

The state-space of $\SpecMDP$ is generated by the outer product of the state-spaces of the environment MDP $\envMDP$ and the reward machine $\SpecFSM$

Thus, the MDP equivalent of a PUnS problem, $\SpecMDP$ generates a problem definition compatible with reinforcement learning algorithms. (In this paper, we utilize discrete representations for state and action spaces; therefore, we use tabular Q-learning \cite{watkins1992q} to compute the policy.)

\subsection{Determining the query execution}
\label{ss:query}
In an active learning framework, the learner generates a query that the teacher must answer by providing a label. There are Many strategies for generating an informative query \cite{settles2009active} have been proposed in prior research. Our strategy is based on the uncertainty sampling approach \cite{lewis1994heterogeneous}, wherein a learner queries about the instance it is least certain how to label . The following is an illustrative example that describes the nature of an informative query selected on the basis of uncertainty sampling.

Consider the table-setting example depicted in Figure \ref{fig:example}. Uncertainty over whether $\varphi_1$ or $\varphi_2$ is the true formula results in a policy that favours  $\varphi_1$, as it is the more restrictive of the two: if the plate and bowl were placed in that order, it would satisfy the specification of just placing the bowl. However, if $\varphi_2$ were the ground truth formula (only the bowl must be placed) , the learned policy would be detrimental to the flexibility of the system during task execution; therefore, it is desirable to refine the belief according to the teacher's feedback.


A task execution where either the fork is placed or both the plate and bowl are placed in that order is not informative; both formulas would label the execution unacceptable or acceptable, respectively. An informative query would attempt to reach the state $\langle \bot, \G \neg Fork \rangle$ by placing only the bowl and not the fork. If this task execution were labeled acceptable, then $\varphi_2$ would be more likely to be the ground truth specification; conversely, if this task execution were judged unacceptable, then $\varphi_1$ would be more likely to be the ground truth specification.

The principle of uncertainty sampling \cite{lewis1994heterogeneous} for active learning states that the most informative query is the one where the current model is most ambivalent about the teacher's expected label. For binary labels, the probability of the query task execution being acceptable should be closest to $0.5$. Given a current belief distribution, $P^{i}(\varphi)$, the learner's estimate of the probability of a trace ($\trace$) being acceptable is computed as follows:

\begin{equation}
\begin{split}
    P(\hat{\mathscr{L}}(\trace) = 1)
    &= \mathds{E}_{P^{i}(\varphi)} \llbracket \mathds{1}(\trace \models \varphi)\rrbracket\\
    &= 0.5\times(1 + R_{\{ \varphi \}}( \langle \varphi' \rangle )
\end{split}
\end{equation}

Here, $\langle \varphi' \rangle$ represents the final state of the reward machine, $\SpecFSM$, after a sequence of transitions described by $\trace$. $P(\hat{\mathscr{L}}(\trace) = 1) = 0.5$ corresponds to a reward value of $0$. Therefore, given a reward machine $\mathscr{M}_{\{\varphi\}}$, the most informative query as per the uncertainty sampling approach should end in a state defined as follows, with $\langle \bm{\varphi} \rangle$ representing the set of terminal states of $\mathscr{M}_{\{\varphi\}}$:

\begin{equation}
    \langle \varphi' \rangle_{selected} = \argmin_{\langle \varphi' \rangle \in \langle \bm{\varphi} \rangle_{term}}
    \mid R_{\{\varphi\}}(\langle \varphi' \rangle) \mid
    \label{eq:select}
\end{equation}

Finally, in order to compute a policy for performing a task execution that terminates in $\langle \varphi \rangle_{selected}$, we reshape the reward values of $\SpecFSM$. Let $\langle \bm{\varphi} \rangle_{path}$ be the set of states that lie along any path joining the initial state, $\langle \varphi\rangle$, and $\langle \varphi' \rangle_{selected}$; the reshaped reward function would then be defined as follows:

\begin{equation}
    R_{shaped}(\langle \varphi' \rangle) = \begin{cases}
    1 &, \langle \varphi' \rangle = \langle \varphi \rangle_{selected} \\
    0 &, \langle \varphi' \rangle \in \langle \bm{\varphi} \rangle_{path} \\
    -1&, \text{ otherwise}
    \end{cases}
\end{equation}

The reshaped reward, $R_{shaped}(\langle \varphi' \rangle)$, is indicated in blue for the dinner table example described in Figure \ref{fig:example}. Note that this reward is only maximized when an execution terminates in $\langle \varphi' \rangle_{selected}$. The policy to generate an informative query execution can be computed by solving the MDP $\mathscr{M}_{spec} = \langle \mathscr{X}\times \FSMStates, \mathscr{A}, T_{spec}, R_{shaped} \rangle$. (Note that this is identical to $\SpecMDP$ apart from reward function.) 

\section{Evaluations}

We evaluated our proposed framework using both a simulated experiment and a user study. The experiment incorporated the synthetic environment proposed in our previous work \cite{shah2018bayesian} to rapidly generate scenarios with varying temporal specifications. We assessed the ability of our proposed framework to infer the correct LTL specifications compared with baselines as described in Section \ref{sec:baselines}, and found that an active learning protocol within our framework generated posterior beliefs that were better aligned with the ground truth specification compared with learning purely from demonstrations or an interactive framework with randomly sampled queries.

\subsection{Baselines}
\label{sec:baselines}

To our knowledge, our proposed framework is the first to model robot learning for non-Markov tasks that unifies demonstrations and a teacher’s acceptability assessments. A natural baseline for our framework is the classical learning-from-demonstrations (LfD) formulation , where the learner learns solely from demonstrations provided by the teacher. We also wanted to evaluate the effect of query selection on learning performance; therefore, as a second baseline, we generated the query executions by selecting actions at each time step from a uniform random distribution. Based on these three paradigms, we used the following three protocols:

\begin{enumerate}
    \item \textbf{Active}: The teacher initially provides two demonstrations, then the learner generates queries. The learner's belief over LTL formulas is updated after an assessment is provided by the teacher  for each of the queries. Each query is generated to reach an informative terminal state, as defined by Equation \ref{eq:select}.
    
    \item \textbf{Random}: This protocol is identical to the \textit{Active} protocol, except that queries are generated by uniformly sampling available actions at each time step.
    
    \item \textbf{Batch}: In this condition, the teacher only provides demonstrations, and the learner can not elicit any assessment of its task performance. The final belief is the posterior distribution computed using Bayesian specification inference \cite{shah2019planning}. 
\end{enumerate}

In each training protocol, the task policy was computed using the final belief compiled with the \textit{minimum regret} criterion. The number of task executions provided to the learner (as either demonstrations or queries) was equal in all cases.

\subsection{Simulation Experiments}
\label{sec:sims}
The task environment for all simulations was based on the synthetic domain \cite{shah2018bayesian}. This domain allows a variable number of threats and waypoints, where the admissible orders for visiting waypoints are encoded within the ground truth formula LTL formula . We allowed  a maximum of five waypoints and five threats for any simulation run; the available action space enabled the learner to select any of these 10 targets to visit.

For all runs of the simulation, the procedure was as follows:
\begin{enumerate}
    \item Select the number of queries $n_{query}$.
    \item A ground truth LTL formula $\varphi$ was sampled from the priors developed in our previous work \cite{shah2018bayesian}.
    \item Two (2) demonstrations that satisfied the ground truth formula were generated and added to the dataset  $\mathscr{D} = \{ \langle \trace_1, 1 \rangle, \langle \trace_2, 1\rangle \}$.
    \item $\mathscr{D}$ was used with the \textit{Active} protocol with $n_{query}$ queries generated by the learner. The final belief, $P_{active}(\varphi)$, was recorded.
    \item $\mathscr{D}$ was used with the \textit{Random} protocol with $n_{query}$ queries generated by the learner. The final belief, $P_{random}(\varphi)$, was recorded.
    \item An augmented dataset, $\mathscr{D}_{batch} = \mathscr{D} \cup \{ \langle \trace_{2+i}, 1 \rangle:  i\in \{1,\ldots, n_{query} \} \}$, was created by generating three additional demonstrations that satisfied the ground truth formula. This dataset was then used with the \textit{Batch} protocol, and the final belief, $P_{batch}(\varphi)$, was recorded. (This ensured the total number of task executions provided to all baselines was equal.)
\end{enumerate}

The experiment was conducted for values of $n_{query} = \{1, \ldots, 6\}$, with 200 runs for each value and a different ground truth formula sampled for each run. For every individual run, the entropy of the final belief and similarity to the ground truth formula were recorded for each of the training protocols. Given two formulas, $\varphi_1$ and $\varphi_2$, that are conjunctive compositions of the clauses in sets $\bm{C}_1$ and $\bm{C}_2$, respectively, the similarity of the two formulas is defined using the  intersection-over-union, as follows:

\begin{equation}
L(\varphi_1, \varphi_2) = \frac{\bm{C}_1 \cap \bm{C}_2}{\bm{C}_1 \cup \bm{C}_2}
    \label{eq:sim}
\end{equation}

The similarity of belief distribution $P(\varphi)$ with ground truth formula $\varphi^*$ is computed as follows:

\begin{equation}
    L(P(\varphi)) = \mathds{E}_{P(\varphi)} \llbracket L(\varphi, \varphi^*) \rrbracket
\end{equation}

$\bm{C}_1$ and $\bm{C}_2$ represent the sets of conjunctive clauses. 

\subsubsection{\textbf{Results}}

\begin{figure}
\centering
  \begin{subfigure}{0.23\textwidth}
      \centering
      \includegraphics[width = \textwidth]{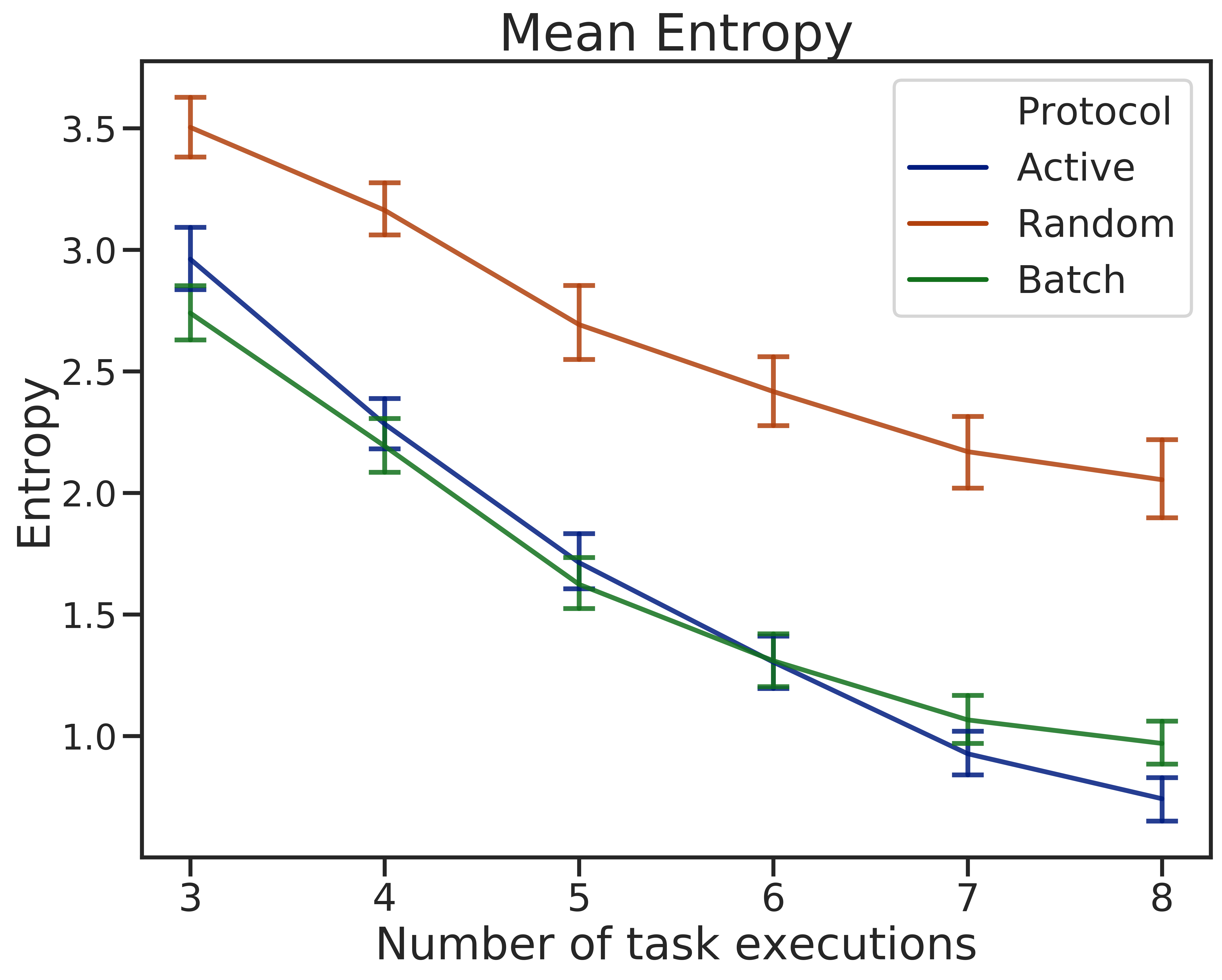}
      \caption{}
      \label{fig:entropy}
    \end{subfigure}
    \begin{subfigure}{0.23\textwidth}
      \centering
      \includegraphics[width = \textwidth]{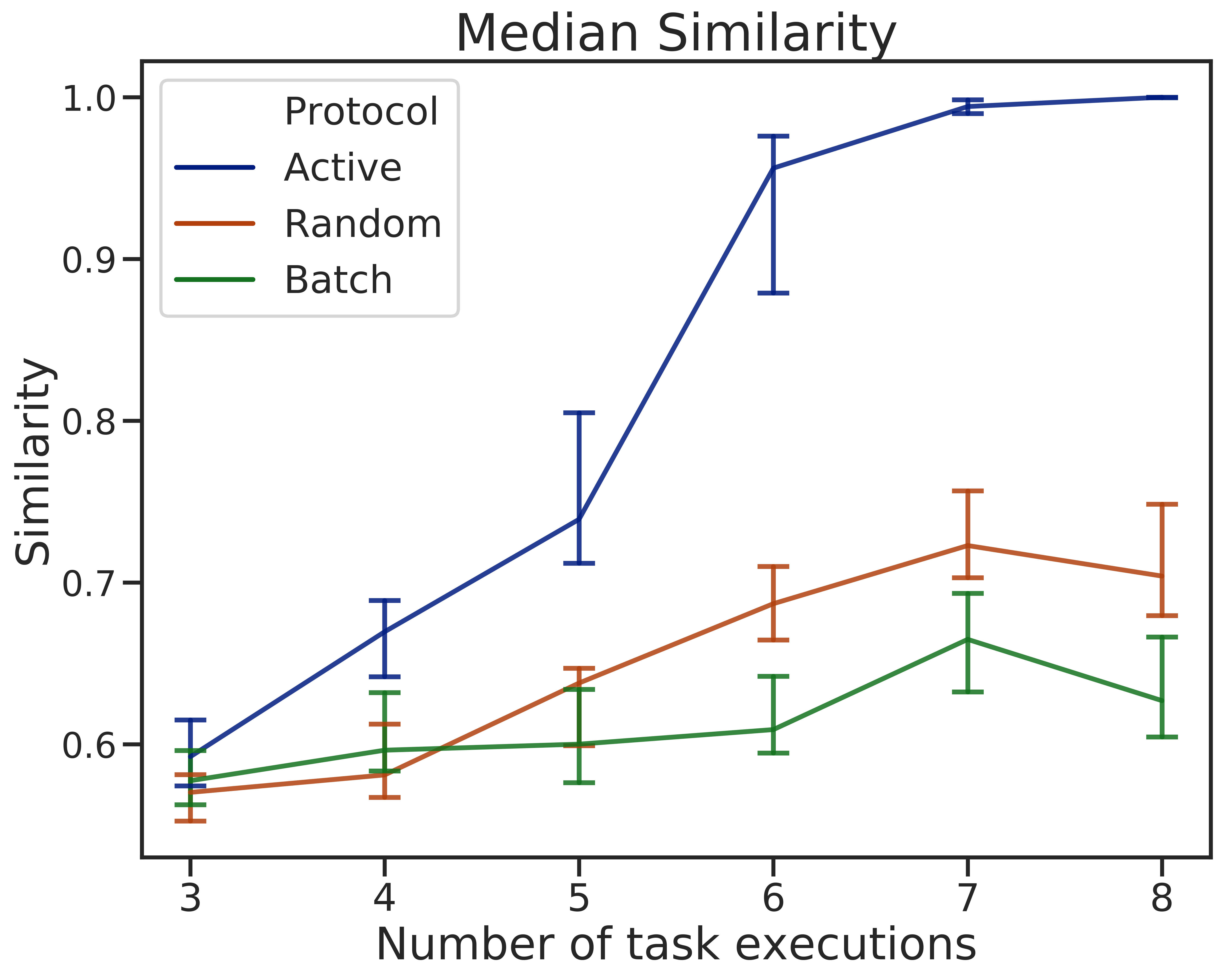}
      \caption{}
      \label{fig:similarity}
    \end{subfigure}
    \caption{The average entropy (left; lower is better) of the final posterior and the similarity of the posterior to the ground truth formula (right; higher is better) for the four training conditions. All error bars indicate 95\% confidence interval.}
    \vspace{-15 pt}
    \label{fig:results}
\end{figure}

\begin{figure*}[h]
\centering
  \begin{subfigure}{0.23\textwidth}
      \centering
      \includegraphics[width = \textwidth]{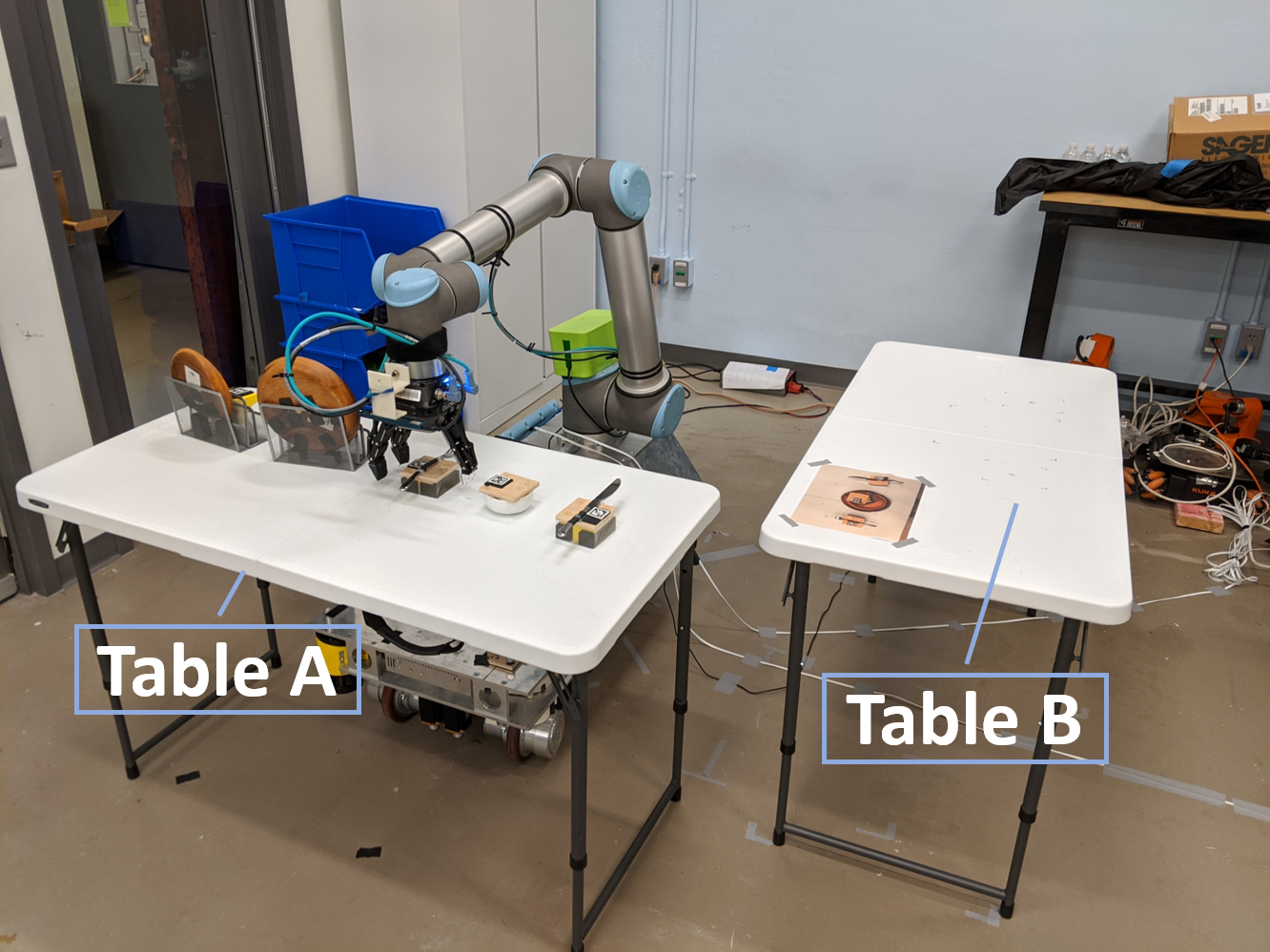}
      \caption{}
      \label{fig:setup}
    \end{subfigure}
    \begin{subfigure}{0.23\textwidth}
      \centering
      \includegraphics[width = \textwidth]{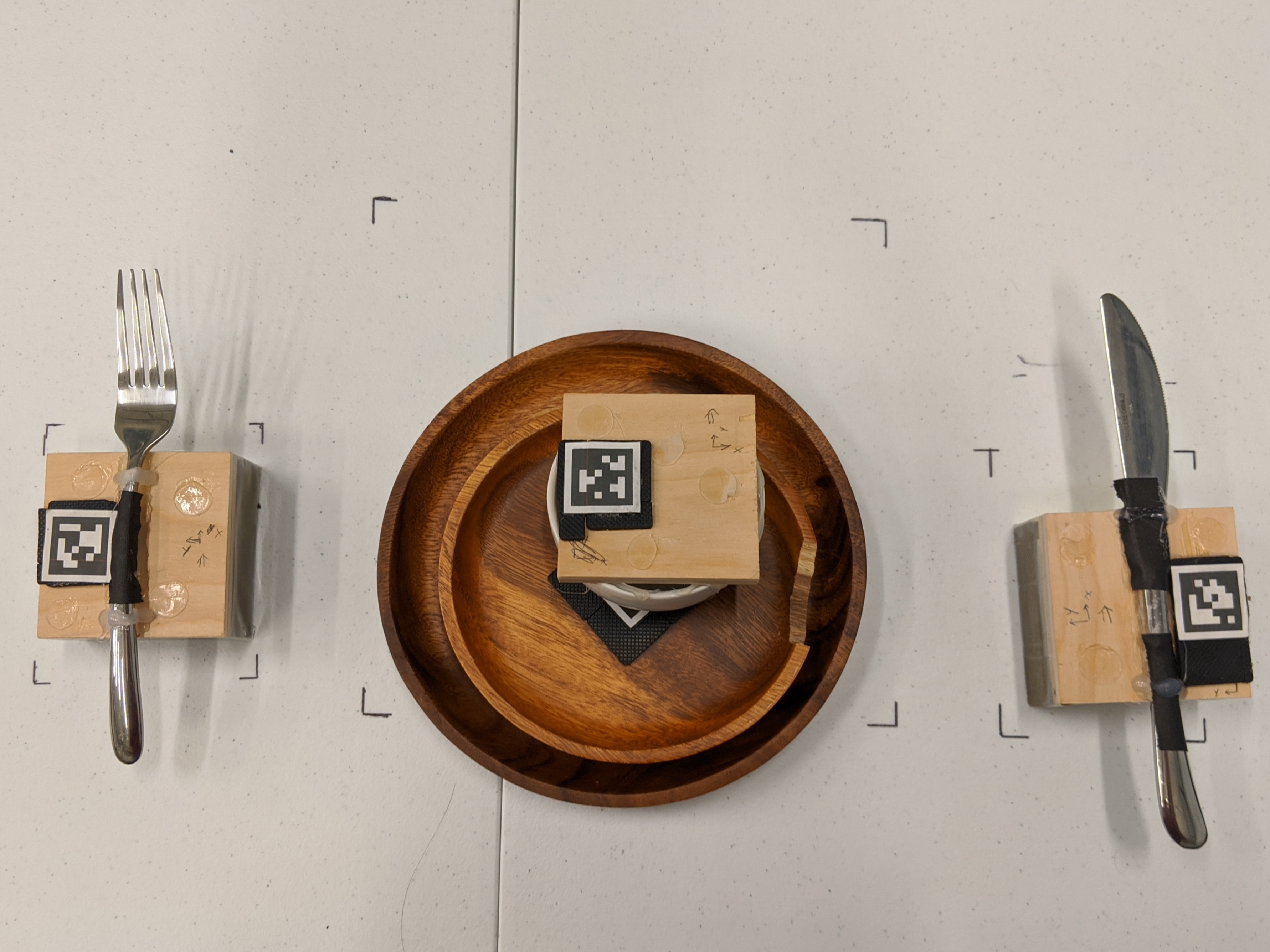}
      \caption{}
      \label{fig:desired}
    \end{subfigure}
    \begin{subfigure}{0.23\textwidth}
      \centering
      \includegraphics[width = \textwidth]{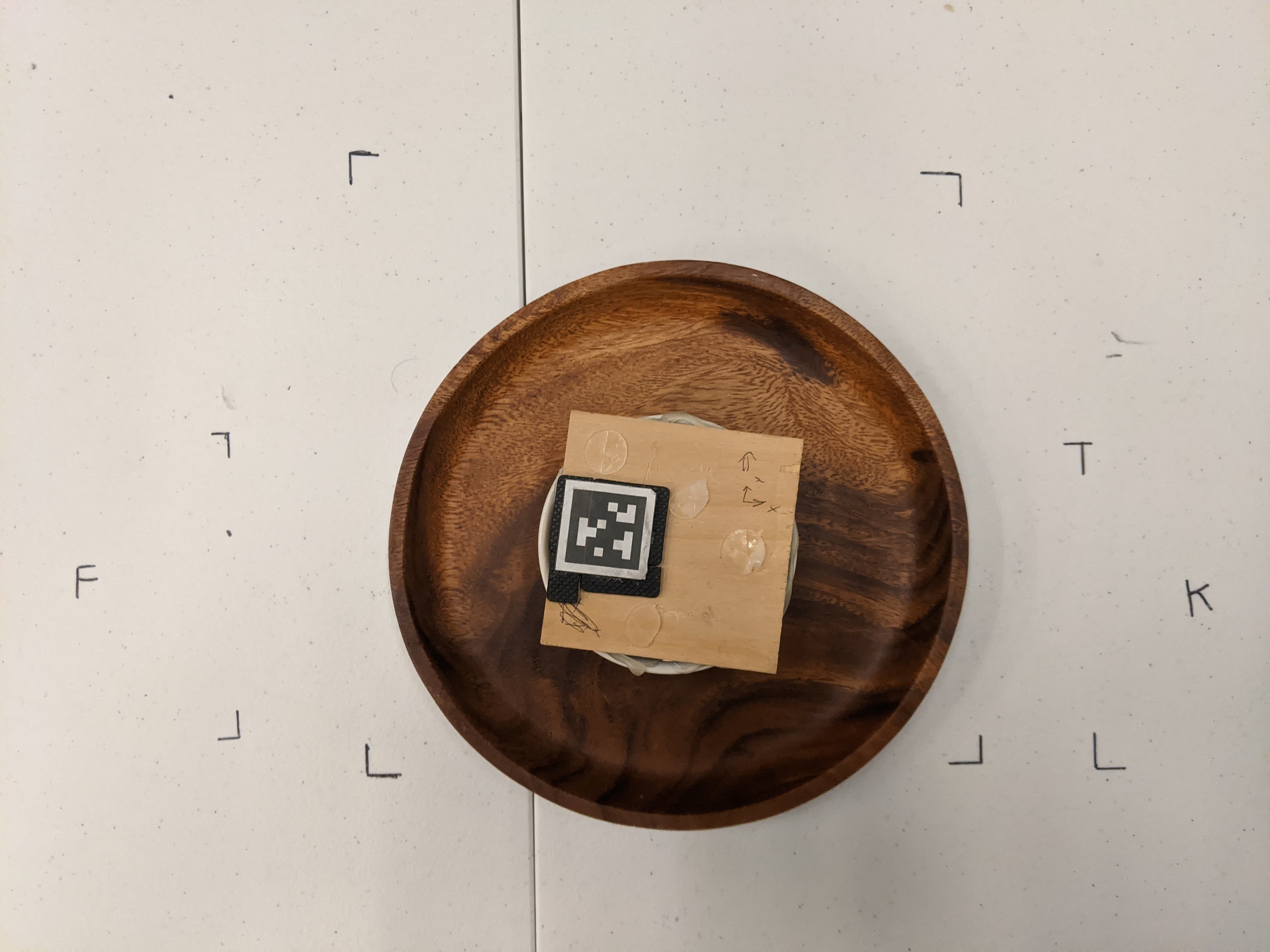}
      \caption{}
      \label{fig:desired2}
    \end{subfigure}
    \begin{subfigure}{0.23\textwidth}
      \centering
      \includegraphics[width = \textwidth]{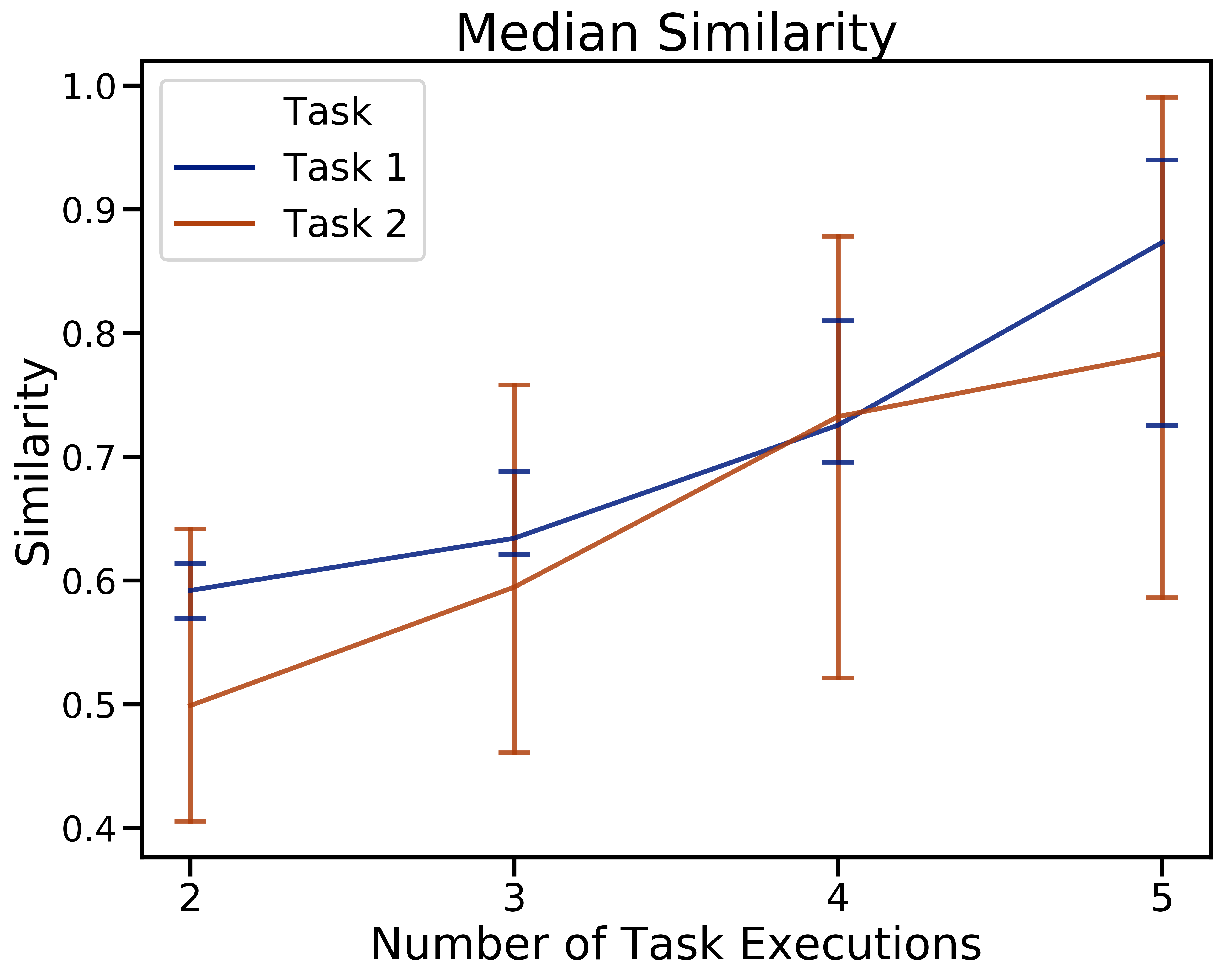}
      \caption{}
      \label{fig:expsimilarity}
    \end{subfigure}
    \caption{Figure \ref{fig:setup} depicts the experiment setup, where objects must be arranged on Table B. Figure \ref{fig:desired} depicts the desired final configuration for \textit{Task 1}, with all objects placed on the table. Figure \ref{fig:desired2} depicts one of the final configurations for \textit{Task 2}, where only the required objects (dinner plate and bowl) are placed, while optional objects (knife and fork) and the prohibited object (small plate) are not.}
    \vspace{-15 pt}
    \label{fig:user}
\end{figure*}

Figure \ref{fig:results} depicts the results from our simulation experiment, and Figure \ref{fig:entropy} depicts the mean entropy value of the final belief for all baselines across all runs. Our results indicate that belief distribution’s entropy decreased as the training protocols processed more labeled task executions; however, this decrease was slower for the \textit{Random} protocol than for the \textit{Active} and \textit{Batch} protocols. This is to be expected, as demonstrations generated through random actions are less informative than either correct demonstrations or the most uncertain task execution (as per the learner's initial belief). Our findings also indicate that both the \textit{Batch} and \textit{Active} protocols yielded similar entropy values, suggesting a similar degree of confidence over the final belief distribution.

Figure \ref{fig:similarity} depicts the median value of the similarity between the final belief and the ground truth formula. The maximum value for the similarity metric is 1, while the minimum is 0. The \textit{Active} protocol outperformed the \textit{Batch} and \textit{Random} protocols with regard to inferring a belief aligned with the ground truth specification. Also, the difference between the median similarity metrics increased with the total number of task executions processed by the training protocols. 

Finally, the low similarity score and entropy observed for the \textit{Batch} protocol indicate that it is susceptible to inferring a belief distribution that is not aligned with the ground truth formula with a high degree of confidence. One potential explanation for this finding is confirmation bias, as multiple identical demonstrations would cause the inference model to assign high probability to an over-constrained formula satisfied by the demonstration. Notably, the similarity of the final posterior learned with the \textit{Active} protocol increases monotonically as additional task executions are provided and performance variance decreases, indicating good  learning performance on all ground truth specifications rather than just a subset thereof. 

\subsection{User Study}
In order to evaluate the real-world performance of the active learning approach, we conducted a user study that involved participants teaching a robot to set a dinner table in various reference configurations. 

Figure \ref{fig:setup} depicts the experiment setup. During each task execution (whether demonstrated by the participant or performed by the robot), the five objects  were initially placed on Table A, and subsequently arranged on Table B. In the first phase of the study, the task (\textit{Task 1}) involved arranging the objects into the configuration depicted in Figure \ref{fig:desired}, with demonstrations provided directly by participants. In the second phase, we modified the study to be conducted online due to the restrictions on in-person studies imposed in light of the COVID-19 pandemic, and participants remotely commanded the robot to provide the demonstrations. The robot's task execution and the experiment instructions were displayed to the participants via video conferencing. In addition to \textit{Task 1}, we also added a second task (\textit{Task 2}), wherein the dinner plate and bowl were to be placed on the table in the configuration depicted in Figure \ref{fig:desired2}. Placing the fork and knife were optional, and placing the small plate was not permitted. (example video: \url{ral2021.ajshah.info}) 

We instructed participants to move only a single object at a time while providing demonstrations, and informed them  that objects could not be picked up again once placed on Table B. Participants were also instructed to provide an assessment after observing the robot while it executed the task; a participant's label was only recorded once the entire task had been completed. For both the in-person and remote study protocols, the participant initiated the robot's belief with  two demonstrations; beliefs were then refined using three queries generated via our active learning models. Finally, the robot demonstrated the results of its learning  by performing three task executions observed by the participant.

The state space of the robot, $\mathscr{X}$, was identical to the set of propositions required for evaluating the task, $\bm{\alpha}$, and contained five Boolean propositions, each of which encoded whether a particular object was successfully placed on the table. The robot's action space, $\mathscr{A}$, comprised five actions (one for each object). Initiating an action triggered a sequence of parameterized primitives programmed into the robot to locate, pick up, and place the object on Table B. Based on the constraints provided to the participants and the robot's action space, the only way to successfully complete \textit{Task 1} was to ensure that the dinner plate, small plate, and bowl were placed in that specific order (the fork and the knife could be placed at any instant). There were multiple final acceptable configurations for \textit{Task 2}; in each, the dinner plate and bowl were placed in that partial order , while the fork and the knife may or may not have been placed.

\subsubsection{Results and Discussions}:

    

We recruited 18 participants for the in-person phase of the study, but had to terminate the protocol with three participants due to robot hardware failure. The results include data collected from 15 participants (10 male, 5 female, median age: 26 years), seven of whom reported prior experience with robots or automated systems. All participants were instructed to teach the robot to perform \textit{Task 1}. For the remote phase, we recruited 12 participants (8 male, 4 female, median age: 28 years); four participants reported prior experience with robotics. We assigned six participants each to \textit{Task 1} and \textit{Task 2}. 

All participants were successfully able to teach the assigned task to the robot — i.e., the policies learned by the robot did not result in an incorrect table setting during any of the test executions. The learning curves for the robot are depicted in Figure \ref{fig:expsimilarity}. 

Overall, the median similarity of the final belief with respect to the ground truth formula was $0.83~ 95 \%~CI: ~[0.73, 0.94]$. For Task 1, the median similarity was $0.87 ~ 95\%~CI:~[0.73, 0.94]$; for Task 2, it was $0.78 ~95\%~CI:~ [0.59, 0.99]$. For \textit{Task 1}, the posterior belief distribution recovered the ground truth formula as the most likely LTL specification for 14 out of 21 participants, while the most likely specification differed from the ground truth by a single conjunctive clause in four cases. Similarly, for \textit{Task 2}, the posterior belief distribution for two out of six participants recovered the ground truth formula as the most likely LTL specification, while the most likely specification for two of them differed from the ground truth formula by a single conjunctive clause. Our demonstration of the entire learning pipeline on an embodied robot indicates the viability of deploying our active learning framework for real-world applications.

\vspace{-5pt}
\section{Conclusion}

Our proposed interactive training framework provides a unified formulation capable of learning non-Markov tasks from both demonstrations provided by a teacher and that teacher's assessment of the robot's task executions. We further proposed an active querying algorithm that allows the learner to identify and perform a task execution with the most uncertain degree of acceptability based on the principle of uncertainty sampling. Finally, we demonstrated the efficacy of our active learning framework for learning non-Markov tasks with a wide range of ground truth specifications through both a simulation experiment and a user study. Notably, the robot performed its task without errors, and the final belief of the robot was closely aligned with the true task specifications across all participants.


\bibliographystyle{ieeetr}
\bibliography{refs}

\begin{thebibliography}{10}

\bibitem{pnueli1977temporal}
A.~Pnueli, ``The temporal logic of programs,'' in {\em Foundations of Computer
  Science, 1977., 18th Annual Symposium on}, pp.~46--57, IEEE, 1977.

\bibitem{shah2018bayesian}
A.~Shah, P.~Kamath, J.~A. Shah, and S.~Li,
  ``\href{http://papers.nips.cc/paper/7637-bayesian-inference-of-temporal-task-specifications-from-demonstrations.pdf}{Bayesian
  Inference of Temporal Task Specifications from Demonstrations},'' in {\em
  Advances in Neural Information Processing Systems 31}, pp.~3804--3813, 2018.

\bibitem{kim2019ijcai}
J.~Kim, C.~Muise, A.~Shah, S.~Agarwal, and J.~Shah, ``Bayesian inference of
  linear temporal logic specifications for contrastive explanations,'' in {\em
  IJCAI}, 2019.

\bibitem{oh2019planning}
Y.~Oh, R.~Patel, T.~Nguyen, B.~Huang, E.~Pavlick, and S.~Tellex, ``Planning
  with state abstractions for non-{Markovian} task specifications,'' in {\em
  RSS}, June 2019.

\bibitem{gopalan2018sequence}
N.~Gopalan, D.~Arumugam, L.~L. Wong, and S.~Tellex, ``Sequence-to-sequence
  language grounding of non-markovian task specifications.,'' in {\em Robotics:
  Science and Systems}, 2018.

\bibitem{shah2019planning}
A.~Shah, S.~Li, and J.~Shah, ``Planning with uncertain specifications
  {(PUnS)},'' {\em IEEE Robotics and Automation Letters}, 2020.

\bibitem{argall2009survey}
B.~D. Argall, S.~Chernova, M.~Veloso, and B.~Browning, ``A survey of robot
  learning from demonstration,'' {\em Robotics and autonomous systems},
  vol.~57, no.~5, pp.~469--483, 2009.

\bibitem{chernova2014robot}
S.~Chernova and A.~L. Thomaz, ``Robot learning from human teachers,'' {\em
  Synthesis Lectures on Artificial Intelligence and Machine Learning}, vol.~8,
  no.~3, pp.~1--121, 2014.

\bibitem{luketina2019}
J.~Luketina, N.~Nardelli, G.~Farquhar, J.~Foerster, J.~Andreas,
  E.~Grefenstette, S.~Whiteson, and T.~Rocktäschel, ``A survey of
  reinforcement learning informed by natural language,'' in {\em Proceedings of
  the Twenty-Eighth International Joint Conference on Artificial Intelligence,
  {IJCAI-19}}, pp.~6309--6317, International Joint Conferences on Artificial
  Intelligence Organization, 7 2019.

\bibitem{bajcsy2017learning}
A.~Bajcsy, D.~P. Losey, M.~K. O’Malley, and A.~D. Dragan, ``Learning robot
  objectives from physical human interaction,'' {\em Proceedings of Machine
  Learning Research}, vol.~78, pp.~217--226, 2017.

\bibitem{bajcsy2018}
A.~Bajcsy, D.~P. Losey, M.~K. O’Malley, and A.~D. Dragan, ``Learning from
  physical human corrections, one feature at a time,'' in {\em Proceedings of
  the 2018 ACM/IEEE International Conference on Human-Robot Interaction}, HRI
  ’18, (New York, NY, USA), p.~141–149, Association for Computing
  Machinery, 2018.

\bibitem{dorsa2017active}
D.~Sadigh, A.~D. Dragan, S.~Sastry, and S.~A. Seshia, ``Active preference-based
  learning of reward functions,'' in {\em Robotics: Science and Systems (RSS)},
  2017.

\bibitem{biyik2018batch}
E.~Biyik and D.~Sadigh, ``Batch active preference-based learning of reward
  functions,'' in {\em Conference on Robot Learning}, pp.~519--528, 2018.

\bibitem{biyik2019asking}
E.~Biyik, M.~Palan, N.~C. Landolfi, D.~P. Losey, and D.~Sadigh, ``Asking easy
  questions: A user-friendly approach to active reward learning,'' in {\em 3rd
  Conference on Robot Learning (CoRL)}, October 2019.

\bibitem{vazquez2018}
M.~Vazquez-Chanlatte, S.~Jha, A.~Tiwari, M.~K. Ho, and S.~Seshia, ``Learning
  task specifications from demonstrations,'' in {\em Advances in Neural
  Information Processing Systems 31}, pp.~5368--5378, 2018.

\bibitem{kasenberg2017interpretable}
D.~Kasenberg and M.~Scheutz, ``Interpretable apprenticeship learning with
  temporal logic specifications,'' in {\em IEEE Conference on Decision and
  Control}, 2017.

\bibitem{camacho2019learning}
A.~Camacho and S.~A. McIlraith, ``Learning interpretable models in linear
  temporal logic,'' in {\em Proceedings of the Twenty-Nineth International
  Conference on Automated Planning and Scheduling ({ICAPS})}, pp.~621--630,
  2019.

\bibitem{cakmak2012designing}
M.~Cakmak and A.~L. Thomaz, ``Designing robot learners that ask good
  questions,'' in {\em Proceedings of the seventh annual ACM/IEEE international
  conference on Human-Robot Interaction}, pp.~17--24, ACM, 2012.

\bibitem{cakmak2010designing}
M.~Cakmak, C.~Chao, and A.~L. Thomaz, ``Designing interactions for robot active
  learners,'' {\em IEEE Transactions on Autonomous Mental Development}, vol.~2,
  no.~2, pp.~108--118, 2010.

\bibitem{cui2018active}
Y.~Cui and S.~Niekum, ``Active reward learning from critiques,'' in {\em 2018
  IEEE International Conference on Robotics and Automation (ICRA)},
  pp.~6907--6914, IEEE, 2018.

\bibitem{kress2009temporal}
H.~Kress-Gazit, G.~E. Fainekos, and G.~J. Pappas, ``Temporal-logic-based
  reactive mission and motion planning,'' {\em IEEE transactions on robotics},
  vol.~25, no.~6, pp.~1370--1381, 2009.

\bibitem{camacho2018finite}
A.~Camacho, J.~A. Baier, C.~Muise, and S.~A. McIlraith, ``Finite {LTL}
  synthesis as planning,'' in {\em Twenty-Eighth International Conference on
  Automated Planning and Scheduling}, 2018.

\bibitem{camacho2019strong}
A.~Camacho and S.~A. McIlraith, ``Strong fully observable non-deterministic
  planning with {LTL} and {LTL-f} goals,'' in {\em Proceedings of the
  Twenty-Eighth International Joint Conference on Artificial Intelligence
  ({IJCAI})}, pp.~5523--5531, 2019.

\bibitem{littman2017environment}
M.~L. Littman, U.~Topcu, J.~Fu, C.~Isbell, M.~Wen, and J.~MacGlashan,
  ``Environment-independent task specifications via {GLTL},'' {\em arXiv
  preprint arXiv:1704.04341}, 2017.

\bibitem{toro2018teaching}
R.~Toro~Icarte, T.~Q. Klassen, R.~Valenzano, and S.~A. McIlraith, ``Teaching
  multiple tasks to an {RL} agent using {LTL},'' in {\em Proceedings of the
  17th International Conference on Autonomous Agents and MultiAgent Systems},
  pp.~452--461, International Foundation for Autonomous Agents and Multiagent
  Systems, 2018.

\bibitem{icarte2018using}
R.~T. Icarte, T.~Klassen, R.~Valenzano, and S.~McIlraith, ``Using reward
  machines for high-level task specification and decomposition in reinforcement
  learning,'' in {\em International Conference on Machine Learning},
  pp.~2112--2121, 2018.

\bibitem{camacho2019ltl}
A.~Camacho, R.~T. Icarte, T.~Q. Klassen, R.~Valenzano, and S.~A. McIlraith,
  ``{LTL} and beyond: Formal languages for reward function specification in
  reinforcement learning,'' in {\em Proceedings of the 28th International Joint
  Conference on Artificial Intelligence (IJCAI)}, pp.~6065--6073, 2019.

\bibitem{bacchus2000using}
F.~Bacchus and F.~Kabanza, ``Using temporal logics to express search control
  knowledge for planning,'' {\em Artificial intelligence}, vol.~116, no.~1-2,
  pp.~123--191, 2000.

\bibitem{manna1987hierarchy}
Z.~Manna and A.~Pnueli, {\em A hierarchy of temporal properties}.
\newblock Department of Computer Science, 1987.

\bibitem{dwyer1999patterns}
M.~B. Dwyer, G.~S. Avrunin, and J.~C. Corbett, ``Patterns in property
  specifications for finite-state verification,'' in {\em Proceedings of the
  21st international conference on Software engineering}, pp.~411--420, ACM,
  1999.

\bibitem{tenenbaum2000rules}
J.~B. Tenenbaum, ``Rules and similarity in concept learning,'' in {\em Advances
  in neural information processing systems}, pp.~59--65, 2000.

\bibitem{laplace1951philosophical}
P.~S. Laplace and P.~Simon, ``A philosophical essay on probabilities,
  translated from the 6th french edition by frederick wilson truscott and
  frederick lincoln emory,'' 1951.

\bibitem{dippl}
N.~D. Goodman and A.~Stuhlm\"{u}ller, ``{The Design and Implementation of
  Probabilistic Programming Languages}.'' \url{http://dippl.org}, 2014.
\newblock Accessed: 2018-4-9.

\bibitem{watkins1992q}
C.~J. Watkins and P.~Dayan, ``Q-learning,'' {\em Machine learning}, vol.~8,
  no.~3-4, pp.~279--292, 1992.

\bibitem{settles2009active}
B.~Settles, ``Active learning literature survey,'' tech. rep., University of
  Wisconsin-Madison Department of Computer Sciences, 2009.

\bibitem{lewis1994heterogeneous}
D.~D. Lewis and J.~Catlett, ``Heterogeneous uncertainty sampling for supervised
  learning,'' in {\em Machine learning proceedings 1994}, pp.~148--156,
  Elsevier, 1994.

\end{thebibliography}
\end{document}